# Context-aware deep learning using individualized prior information reduces false positives in disease risk prediction and longitudinal health assessment

Lavanya Umapathy[1,2], Patricia M Johnson[1,2,5], Tarun Dutt[1,2], Angela Tong[1], Madhur Nayan[3,4], Hersh Chandarana[1,2,5], and Daniel K Sodickson[1,2,5]

[1]Bernard and Irene Schwartz Center for Biomedical Imaging, Department of Radiology, New York University Grossman School of Medicine, New York, NY, USA, [2]Center for Advanced Imaging Innovation and Research (CAI²R), Department of Radiology, New York University Grossman School of Medicine, New York, NY, USA, [3]Department of Urology, New York University Grossman School of Medicine, New York, NY, USA, [4]Department of Population Health, New York University Grossman School of Medicine, New York, NY, USA, [5]Vilcek Institute of Graduate Biomedical Sciences, New York University Grossman School of Medicine, New York, NY, USA

**ABSTRACT**

Temporal context in medicine is valuable in assessing key changes in patient health over time. We developed a machine learning framework to integrate diverse context from prior visits to improve health monitoring, especially when prior visits are limited and their frequency is variable. Our model first estimates initial risk of disease using medical data from the most recent patient visit, then refines this assessment using information digested from previously collected imaging and/or clinical biomarkers. We applied our framework to prostate cancer (PCa) risk prediction using data from a large population (28,342 patients, 39,013 magnetic resonance imaging scans, 68,931 blood tests) collected over nearly a decade. For predictions of the risk of clinically significant PCa at the time of the visit, integrating prior context directly converted false positives to true negatives, increasing overall specificity while preserving high sensitivity. False positive rates were reduced progressively from 51% to 33% when integrating information from up to three prior imaging examinations, as compared to using data from a single visit, and were further reduced to 24% when also including additional context from prior clinical data.  For predicting the risk of PCa within five years of the visit, incorporating prior context reduced false positive rates still further (64% to 9%). Our findings show that information collected over time provides relevant context to enhance the specificity of medical risk prediction. For a wide range of progressive conditions, sufficient reduction of false positive rates using context could offer a pathway to expand longitudinal health monitoring programs to large populations with comparatively low baseline risk of disease, leading to earlier detection and improved health outcomes.

**Keywords:** medical risk prediction; longitudinal monitoring; representation learning

# INTRODUCTION

Healthcare today is largely episodic. Patients typically enter the healthcare system in response to particular symptoms or other signs of a disease, launching a cascade of diagnostic and therapeutic activities which constitute a particular episode of care. Between such intensive episodes, interactions with healthcare providers tend to be comparatively sparse. This punctuated approach to health, together with an intensive medical workload, prioritizes attending to a patient's current needs and making sense of a flood of incoming health measurement data over routine assessment of potentially subtle trends over time. Nevertheless, temporal information is extremely helpful in assessing key changes which may signal the onset or progression of disease. In fields like remote sensing, where multiple variables are monitored at frequent intervals, time series analysis has been used effectively to identify noteworthy trends [1].  By contrast, observations in the field of medicine, especially those associated with lower-risk populations, often have limited longitudinal data, with variable time intervals between visits, and with different data (blood tests, imaging exams, biopsies, etc.) collected during different visits. In any given interaction with a patient, human physicians still use whatever longitudinal information they can to identify trends and look for changes in that patient's baseline state of health, subject to limitations on both the quality/consistency of data and the capacity of the unaided human mind for principled prediction.

Recent years have seen an increasing focus on personalized, preventative health. For example, screening and surveillance for diseases such as prostate, breast, and lung cancer have gained traction in medical practice as a means of promoting early detection [2-8] . Such proactive screening approaches are known to carry a risk of overdiagnosis and overtreatment as a result of the limited specificity of screening tests [9-11]. The specificity of individual screening tests is not entirely inherent to the tests themselves, however. To some extent, specificity is limited by the fact that these tests are often interpreted in isolation. Bayesian statistics suggest that simply repeating the same screening tests over time can reduce false positive rates, if earlier tests are used to update prior probabilities of disease [12] . Comparing the results of repeated tests, and placing these tests in suitable multimodal context, can also facilitate the accurate identification of concerning changes. However, a rigorous means of identifying such concerning changes and distinguishing them from normal variants is essential.

Building on these insights, we set out to design a machine learning framework that systematically incorporates temporal context to improve risk assessment in health monitoring, particularly when the number of monitoring sessions is comparatively small, and when the intervals between sessions are variable. Our hypothesis was that suitable incorporation of diverse temporal information will drive down false positive rates, increasing specificity without compromising sensitivity.

When considering what models to use, we were guided by observations of how human physicians tend to analyze prior information. Clinicians often assess the most recent snapshot of a patient's health, as represented in clinical data and/or radiologic imaging, for example, to estimate a baseline risk of disease based on population norms. They then use any available prior data (including clinical and/or imaging data) to refine their original risk estimate. If current data displays notable changes from prior data, the initial risk estimate is upgraded. Conversely, for findings that are stable over an extended period, the initial risk estimate may either remain constant or be downgraded. Such an approach is highly flexible, and it inherently incorporates key general knowledge about temporal evolution without placing constraints on temporal sampling. We therefore chose to base our automated risk assessment approach on this model of progressive risk refinement.

To demonstrate the benefits of incorporating prior context into medical risk prediction, we chose the example of prostate cancer (PCa) monitoring in a large patient population (n=28,342) studied over nearly a decade (2015-2023), with variable time intervals between radiologic imaging examinations and sampling of clinical biomarkers. For slow-growing cancers such as PCa, temporal context is particularly relevant. At-risk patients are often followed with repeated Magnetic Resonance Imaging (MRI) and/or blood tests for prostate-specific antigen (PSA) [13-16], in order to detect changes which might signal a progression requiring intervention.

We developed a deep learning-based risk refinement framework that first estimates current and future disease risk using data from the most recent patient visit then refines this assessment using information summarized from previously collected imaging and/or clinical biomarkers. As compared with risk assessments based only on current data, our risk steering approach incorporating prior context directly converts false positives to true negatives, increasing specificity while preserving sensitivity for detecting clinically significant prostate cancer both at the time of the visit and five years following the visit. Such increases in specificity are seen using clinical data alone, using imaging data alone, and using combinations of both clinical and imaging data. Combinations of clinical and imaging data yield the greatest benefits.

While we have evaluated our framework in the context of prostate cancer monitoring, similar approaches could have value for a wide range of diseases (other cancers, cardiovascular disease, neurodegenerative disease, etc.) in which temporal information is collected but is not integrated systematically. Moreover, if false positive rates can be decreased sufficiently using context, longitudinal monitoring can be offered to broader populations at progressively lower risk, resulting in earlier detection and improved health outcomes.

## RESULTS

### Overview of our risk refinement framework

Figure 1 presents a schematic overview of our risk refinement approach applied to predicting the risk of prostate cancer. Our framework consists of a representation learner model, a risk estimation model, and a temporal learner model with risk refinement (Methods).

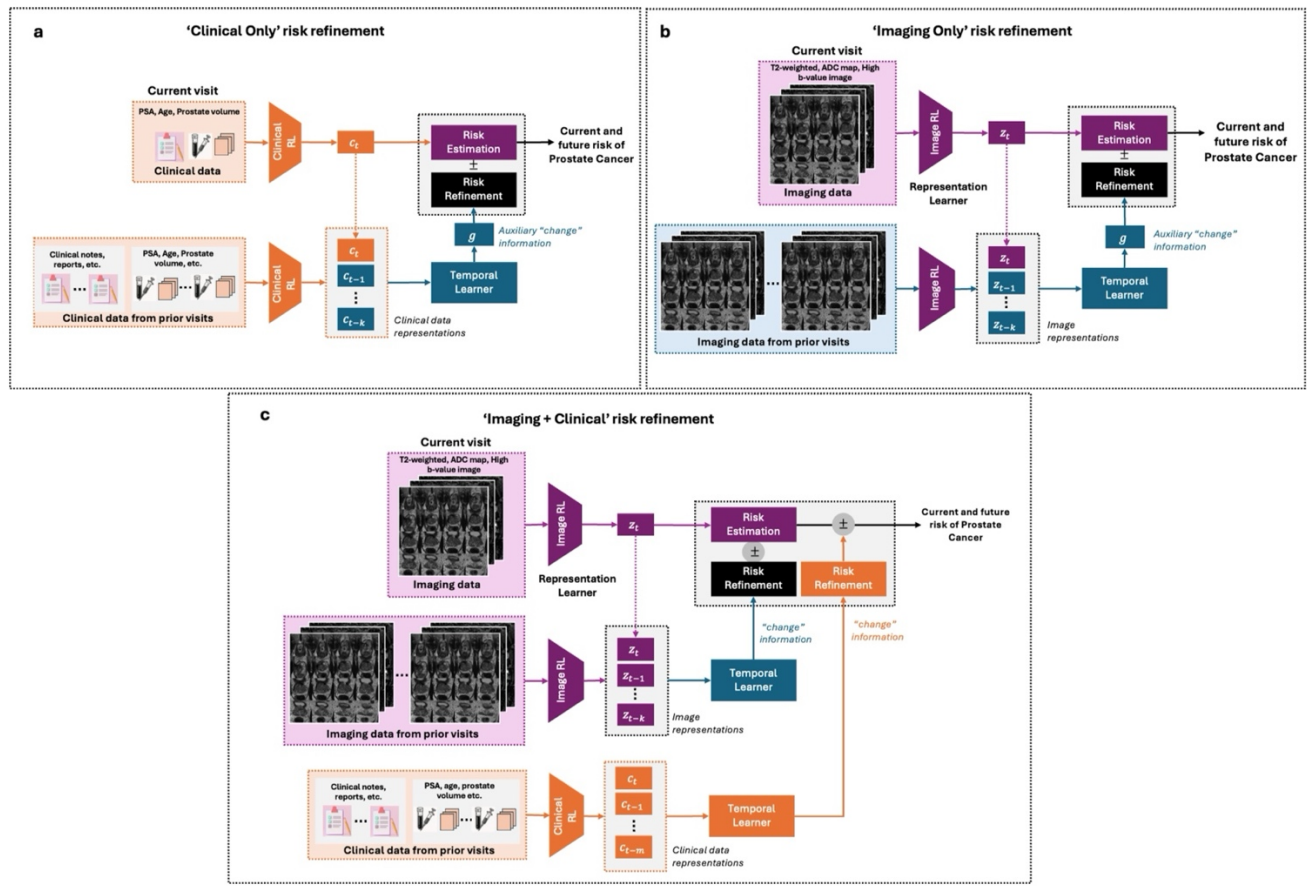

**Figure 1:** Our risk refinement framework uses clinical and/or radiologic imaging data from the most recent patient visit to predict the likelihood of clinically significant prostate cancer at the time of the visit (current) and within 5-years, and leverages patient-specific information from prior visits, whenever available, to refine this initial assessment. This framework consists of a representation learner (RL) model that transforms clinical data or high-dimensional radiologic imaging data into subject-specific low-dimensional latent representations, a risk estimation model that predicts risk of clinically significant prostate cancer based on the most recent representation, and temporal learner (TL) and risk refinement (RR) models that aggregate temporal history (in the form of an arbitrary number of longitudinal representations distilled from previous visits along with corresponding times of those visits with respect to the current visit) into a change signal to steer the initial risk assessment. The TL model aggregates these subject-specific representations from current and prior visits to generate a low-dimensional auxiliary "change" signal, which then guides the risk refinement model to steer our initial risk estimate, i.e., to upgrade or downgrade the initial estimate

to yield a refined risk estimate. This entire framework is trained end-to-end in a supervised fashion (Methods) with longitudinal monitoring data from the large population dataset described below. This temporal risk refinement approach is generalizable to different types of data. Its application to prostate cancer prediction with (a) Clinical Only, (b) Imaging Only, and (c) Combined Imaging + Clinical data are shown here. For the combined risk refinement, the initial risk assessment is updated first using changes based on imaging priors, then using by changes based on clinical priors. Note that the changes in risk over time can also potentially be learned from other types of multi-modal prior information such as clinical variables or textual information from clinician notes or reports.

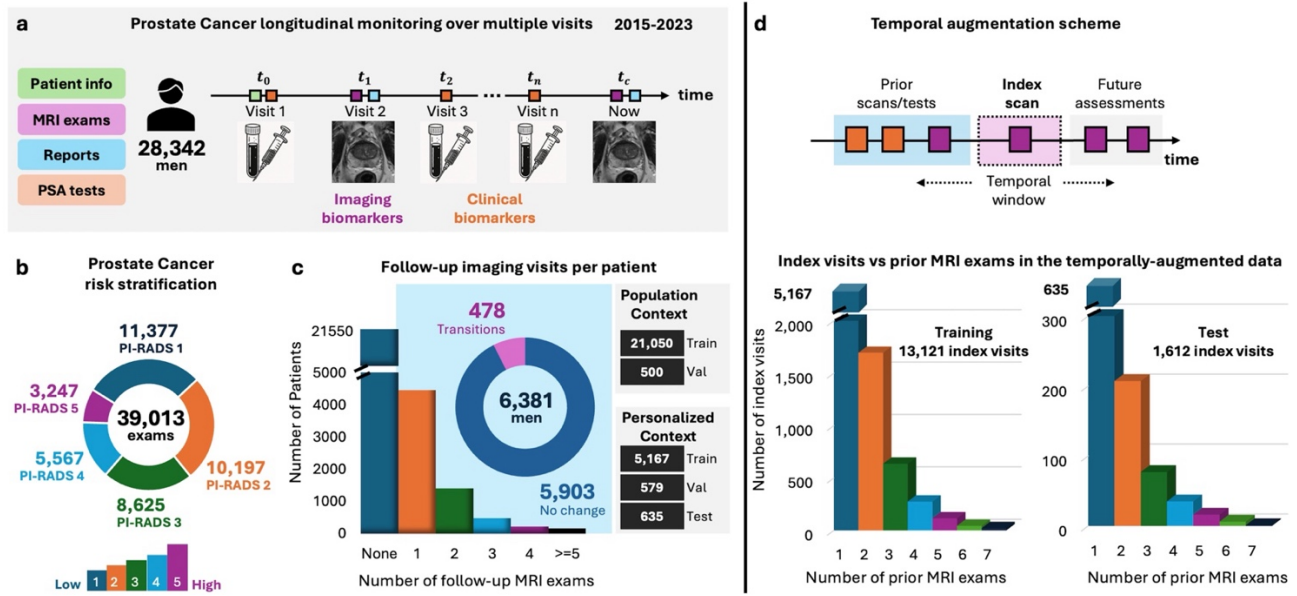

**Figure 2:** Overview of our longitudinal prostate cancer (PCa) dataset. (a) Clinical (68,931 PSA blood tests) and imaging (39,013 MRI exams) biomarkers from longitudinal monitoring of 28,342 men were collected between 2015 and 2023. (b) Breakdown of prostate cancer risk stratification based on maximum PI-RADS assessment by radiologists. In this dataset, 55.3% of exams were deemed no/low-risk for PCa (PI-RADS 1 and 2), 22.1% were deemed intermediate risk for PCa (PI-RADS 3), and 22.6% were deemed high-risk (positive) for PCa (PI-RADS 4 or 5). (c) Bar plot of the number of follow-up imaging visits available for our patient population. 21,550 patients had no follow-up imaging, and 6,792 patients had at least one follow-up MRI, with an average time between initial and follow-up imaging of 1.8±1.1 years. During this study period, 86.9% (n=5,903) of the men with repeated radiologic imaging continued to be in the risk-state assessed during the initial radiologic examination, 7.0% (n=478) transitioned from no/low- or intermediate-risk to high-risk, and 6.0% (n=411) transitioned from high-risk to intermediate/low-risk (see Figure S1 for reasons to exclude this group). Data from 21,550 men with no follow up imaging were used to learn relevant features of prostate cancer risk. 6,381 men with repeated longitudinal monitoring with MRI were used to learn and evaluate risk refinement with prior context. Patients with longitudinal follow-up were randomly split into training (n=5,167; 13,121 MRI exams), validation (n=579; 1,482 MRI exams), and test sets (n=635; 1,612 MRI exams) for training, validation, and testing, respectively, of the AI prostate cancer risk refinement model. (d) Schematic illustration of temporal augmentation around a selected index visit to increase the size of the longitudinal dataset available for risk prediction and evaluation. For each patient with imaging data at multiple timepoints, we randomly selected data from a timepoint of interest as the index visit. All medical data (imaging and clinical) prior to this index visit were considered prior information. Radiologic assessments corresponding to subsequent MR exams were considered future diagnosis. By temporally shifting this selection window, we performed temporal augmentation on our longitudinal imaging data to further increase the size of training and evaluation datasets available to us. (e) Distribution of number of prior exams in the temporally augmented training and test sets. The augmentation procedure resulted in expanded training (n=7,954 index visits with at least one prior MRI), validation (n=903 index visits), and test (n=977 index visits) sets. In the augmented test set, 977 index visits had at least one prior MRI, 342 index visits had at least two prior MRIs, 135 index visits had at least three prior MRIs, and so on.

**Longitudinal prostate data summary**

Our longitudinal prostate cancer dataset (Figure 2, Supplementary-Figure-S1) consisted of data from 28,342 men who underwent bi-parametric MRI of the prostate for known or suspected prostate cancer between 2015 and 2023 at 56 locations in our health enterprise. The mean age of patients in this population was 65.3±8.4 years, mean PSA levels were 7.3 ± 42.9 ng/mL, and mean prostate volumes were 65.8 ± 42.2 cc (Supplementary-Table-S1). Other key characteristics of the dataset, including the risk of clinically significant prostate cancer assessed at the time of imaging with Prostate Imaging Reporting and Data System (PI-RADS) [17 18], the distribution of follow-up studies, and the training/validation/test splits, are also highlighted in Figure 2.

Imaging and clinical data from patients with no follow-up imaging (n=21,550) were used to train our risk refinement model to predict the current risk of PCa based on population context. We used temporally augmented data (Figure 2d) from patients with longitudinal follow-up (personalized context) for end-to-end training and evaluation of our risk refinement framework. Hereafter, all reported results are derived from the temporally augmented training, validation, and test sets.

**AI model evaluation**

We trained three different models utilizing different modalities of data (Methods) to demonstrate the generalizability of our risk refinement approach: a) a 'Clinical Only' risk refinement model that used current and prior clinical data (PSA, age, prostate volume), b) an 'Imaging Only' risk refinement model that used current and prior images, and c) a multi-modal ('Imaging + Clinical') risk refinement model that used current images together with both imaging and clinical priors.

Given clinical data or bi-parametric prostate MRI data from an index visit, together with clinical data and/or MRI data from prior visits, we used our risk refinement models to predict the likelihood of prostate cancer at the time of the index visit (current risk) and within the next five years (5-year risk). In each case, we compared the performance of our risk refinement model incorporating prior context to the performance of baseline risk estimates made from just the index visit without prior context. The bar plots in Figure 3 (plus Supplementary-Tables-S2 and S3) demonstrate the effect of progressively adding prior information on the performance of the different risk refinement models for a subset of the test population (Supplementary-Figure-1c) with at least three prior visits.

***Effect of integrating prior context on clinical data-based risk assessment***: Figure 3a shows the effect of integrating prior information into a risk refinement framework based on longitudinal clinical data on a subset of the test set with at least three prior PSA tests (n=279, positive=44). Here, the clinical data at the index visit consisted of PSA tests performed within 6 months of imaging. With just the clinical data from the index visit, the clinical risk prediction model had a sensitivity of 93% and a specificity of 25% for predicting the current risk of PCa. As we integrated prior context from one, two, and three prior visits using the temporal learner and risk refinement module, the false positive rates dropped from 75% to 62% without affecting true positive rates (93%). This corresponds to increases in specificity from 25% to 38% at constant sensitivity. A similar trend is observed for 5-year risk (n=89, positive=56, false positive rate drops from 73% to 48% at constant true positive rate corresponding to 95% sensitivity).

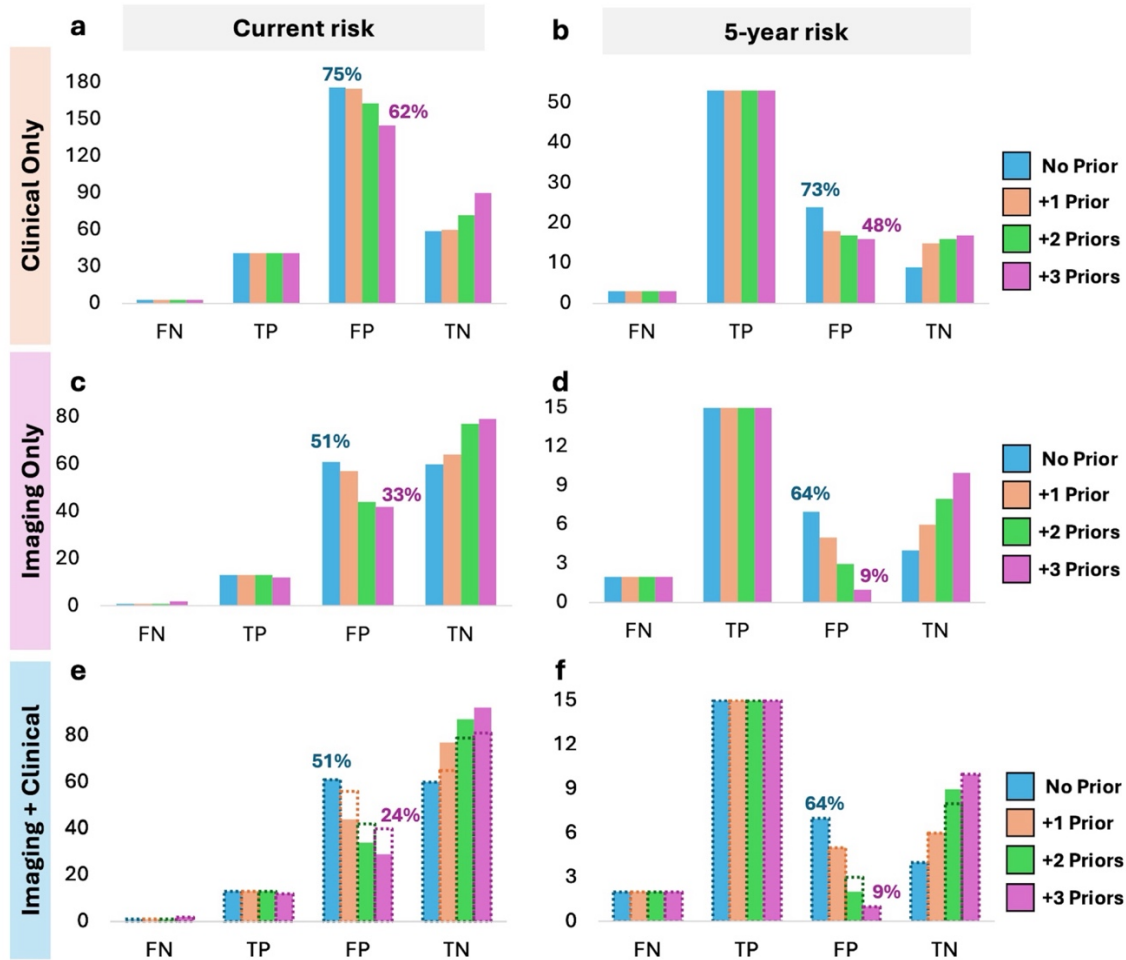

**Figure 3:** Reducing false positive rates in risk prediction models with prior context: Comparison of model performance when predicting current (left) and 5-year (right) risk of prostate cancer on test cases with at least three prior visits. The sensitivity of each model is represented by the balance of false negatives (FN) and true positives (TP), and the specificity of each model is represented by the balance of false positives (FP) and true negatives (TN). Bar charts compare numbers of FN, TP, FP, and TN cases with incorporation of information from 0, 1, 2 and 3 prior visits. For each model, false positive rates (%) are indicated above the false positive bars for no prior and 3 priors. Note that the total number of cases is smaller for 5-year risk than for current risk, as we only used the subset of the test cases with outcome assessments five years after the index visit. (a-b): The clinical model uses baseline clinical data (prostate-specific antigen blood tests, age, and prostate volume) together with clinical data from up to three prior visits for risk refinement. The three prior visits (Priors 1, 2, and 3) were selected from all PSA tests such that Prior 1 was the most recent visit (average time interval from index visit = 7 ± 6 months), Prior 3 was the initial visit (47 ± 25 months), and Prior 2 was the midpoint (25 ± 14 months). (c-d) The imaging model uses baseline prostate MR imaging data together with imaging data from up to three prior visits for risk refinement. Here, the three prior exams were taken from three consecutive prior imaging visits, with average intervals between index visit and prior visit of 2.1 ± 1.2 years for Prior 1, 3.3 ± 1.3 years for Prior 2, and 4.6 ± 1.4 years for Prior 3. (e-f) The combined multi-modal modal uses baseline imaging data together with clinical priors (whenever available) and imaging exams from up to three prior visits for risk refinement. Here, the three imaging priors (Prior 1, 2, and 3) were selected from three consecutive prior visits as described previously, whereas the clinical priors were from the initial visit and the most recent visit. Here, "+1 Prior" indicates use of index visit and Prior 1 for imaging plus both clinical priors, "+2 Priors" indicates use of index visit and Priors 1-2 for imaging plus both clinical priors, and "+3 Priors" indicates use of index visit and Priors 1-3 for imaging plus both clinical priors. To facilitate visual comparison, dotted lines show the performance of the imaging model using risk refinement with imaging priors only. In each case, integration of temporal context reduces false positives by converting them into true negatives, without compromising model sensitivity (see Supplementary Figure-S2 for details on the one patient that was misclassified as negative with imaging priors).

***Effect of integrating prior context on imaging-based risk assessment***: Figure 3c shows the effect of risk refinement with prior imaging context on a subset of the test set with at least three prior MRI exams (n=135, positive=14). Compared to using just the clinical data, the use of an MRI exam from an index visit increased the specificity from 25% to 49% with a sensitivity of 93% for current risk. Once again, we observed that with the addition of each prior MRI exam, more false positives were converted to true negatives. The index false positive rate of 51% was reduced to 46% with additional context from one recent prior MRI exam (Current exam + Prior 1), to 35% with two recent prior MRI exams (Current + Priors 1-2), and to 33% with three prior MRI exams (Current + Priors 1-3), resulting in a specificity of 67% with three priors. Again, a similar trend is observed for 5-year risk (n=28, positive=17, false positive rate drops from 64% to 9% at constant true positive rate corresponding to 88% sensitivity). Integration of temporal context reduced false positives by converting them into true negatives without compromising model sensitivity, with the exception of one patient that was misclassified as a false negative due to implants that were not present previously (see Supplementary-Figure-S2 for details).

***Effect of integrating prior context on multi-modal risk assessment***: Figure 3e shows the performance of the combined model in which both imaging and clinical data are used to refine risk estimates from an index MRI exam. When clinical data from prior visits were available, we performed successive risk refinement with imaging priors followed by clinical priors (Methods, Figure 1c). Whenever clinical priors were not available, we used only imaging priors so as to maintain a consistent number of cases to compare the effect of incorporating multi-modal prior context (Figure 3e-f) over imaging-only context (Figure 3c-d). This is, however, a realistic situation as not all priors would be available at all times.

The use of combined priors reduced false positive rates for current risk of PCa from 51% to 36%, 28%, and 24% for one, two, and three priors, respectively – an improvement of 10%, 7% and 9%, respectively, over the model using imaging priors only. This corresponds to an increase in specificity from 49% without priors to 76% with all priors, at comparable sensitivity. For 5-year risk, an overall reduction in false positive rate from 64% without priors to 9% with three priors is observed, with an improvement of 9% over imaging priors alone in the two-prior case, and equivalent performance for one and three priors. Note that, given the flexibility of our risk refinement framework, imaging and clinical priors for this combined model need not be derived from the same visits, and the number of imaging priors need not match the number of clinical priors.

A performance comparison of the three models (Clinical Only, Imaging Only, and Imaging + Clinical) for a larger test set with contextual information from just one prior visit is presented in Table 1. We observe similar trends as in Figure 3 with the addition of context from just one prior visit: sensitivity remains essentially constant while specificity increases significantly. Such significant increases in specificity are seen using clinical data alone (McNemar's test, $p<0.001$), using imaging data alone ($p<0.001$), or combining both clinical and imaging data ($p<0.001$), with combinations yielding the greatest benefits. There were no significant changes to model sensitivity.

The Imaging Only risk refinement model also showed excellent agreement between the true positive cases predicted with and those predicted without temporal context, for both the current risk of PCa (Cohen's kappa score $\kappa= 0.88$) and the 5-year risk ($\kappa= 0.80$). This was also true for the Clinical Only risk refinement model, where the predicted true positives with and without temporal context showed excellent agreement for both current risk ($\kappa= 0.9$) and 5-year risk of PCa ($\kappa= 1.0$). These scores highlight the fact that integrating temporal information moves false positives to true negatives while maintaining true positives identified using the most recent data from the index visit.

**Table 1:** Effect of integrating prior context into risk refinement for various prostate cancer risk prediction models. Here, the bold values correspond to significant changes with prior context.

| | Current risk | | | 5-year risk | | |
|---|---|---|---|---|---|---|
| | **Specificity** | **Sensitivity** | **AUC [95% CI]** | **Specificity** | **Sensitivity** | **AUC [95% CI]** |
| **Clinical Only** Index visit + clinical prior | (n=383, positive=62) | | | (n=123, positive=76) | | |
| Index visit only | 0.23 (74/321) | 0.92 (57/62) | 0.79 [0.72, 0.85] | 0.30 (14/47) | 0.93 (71/76) | 0.84 [0.77, 0.92] |
| w/ prior clinical context | **0.36**** (116/321) | 0.90 (56/62) | 0.82 [0.76, 0.88] | **0.49*** (23/47) | 0.93 (71/76) | 0.85 [0.78, 0.92] |
| **Imaging Only** Index visit + imaging prior | (n=977, positive=162) | | | (n=298, positive=185) | | |
| Index visit only | 0.49 (397/815) | 0.91 (148/162) | 0.86 [0.83, 0.90] | 0.55 (62/113) | 0.90 (166/185) | 0.88 [0.84, 0.92] |
| w/ prior imaging context | **0.54**** (442/815) | 0.92 (149/162) | **0.88**‡ [0.85, 0.91] | **0.67*** (76/113) | 0.89 (165/185) | **0.90*** [0.86, 0.93] |
| **Imaging + Clinical** Index visit + multi-modal prior | (n=443, positive=71) | | | (n=138, positive=86) | | |
| Index visit only | 0.43 (162/372) | 0.91 (65/71) | 0.85 [0.80, 0.90] | 0.44 (23/52) | 0.91 (78/86) | 0.84 [0.77, 0.91] |
| w/ prior multi-modal context | **0.58**** (217/372) | 0.90 (64/71) | **0.88**‡ [0.83, 0.93] | **0.61**‡ (32/52) | 0.88 (76/86) | **0.88*** [0.82, 0.94] |

Current risk = likelihood of prostate cancer at the time of the index visit; 5-year risk = likelihood of prostate cancer within 5-years of the index visit; AUC = area-under-receiver-operating-characteristic curves; CI = confidence intervals, 95% CI = 95% confidence intervals, Clinical Data Only = risk refinement with only clinical data (log PSA, age, prostate volume) from index visit and one prior visit (first visit); Imaging Only = risk refinement with only prostate MRI from index visit and one prior visit (recent visit); Imaging + Clinical = risk refinement with prostate MRI from index visit, imaging context from one prior visit (recent visit) and clinical context from the first and the most recent visit. **p<0.001, *p<0.01, ‡p<0.05

**Differential risk refinement with prior context**

On incorporating prior context, risk refinement frameworks steered initial risk estimates from the index visit differentially based on the underlying risk status (Supplementary-Figure-S3) or the timing of the prior visit with respect to the index visit. Figure 4 (Supplementary-Table-S4) compares model performance in predicting current risk of PCa when integrating just one prior exam from three different time periods. We observed that the further the prior visit was separated from the index visit, the more false positives were converted to true negatives. These results, which held true both for Imaging Only and for Clinical Only models, are consistent with the general intuition that risk is lower when features are stable over a longer period.

**Correlation with pathology**

To investigate if the predicted risk also correlated with the presence of clinically significant prostate cancer as confirmed by histopathological evaluation, we evaluated our risk refinement model on a secondary test set consisting of patients (Figure 5, Supplementary-Figure-4) with an imaging visit during 2024 and subsequent biopsies within 6 months of imaging who also had at least three prior imaging visits (n=96). Of the 29 patients with clinically significant prostate cancer identified during biopsy, the risk refinement model identified 24 (sensitivity=83%) as high-risk using the most recent MRI, with a specificity of 42%. As additional context from prior imaging visits was incorporated, the false positive rate of the model was progressively reduced from 58% to 45% (Figure 5b) while maintaining constant sensitivity (Cohen's kappa score κ= 1.0).

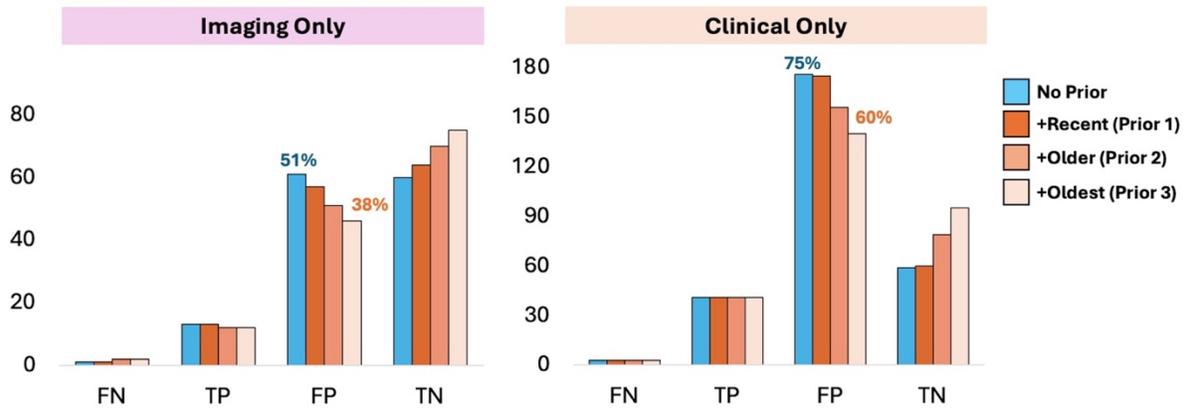

**Figure 4**: Effect of time interval between index visit and prior visit on risk refinement. The sensitivity of each model is represented by the balance of false negatives (FN) and true positives (TP), and the specificity of each model is represented by the balance of false positives (FP) and true negatives (TN). Bar charts compare numbers of FN, TP, FP, and TN cases with incorporation of information from no priors and one prior only (Prior 1, 2, or 3). Integrating data from one prior visit improves specificity differentially depending on how far apart the two visits are in time. For both the Imaging Only and the Clinical Only risk refinement models, using the oldest prior data reduces false positive rate and improves specificity the most. In the Imaging Only risk refinement framework, the inclusion of additional context from just one prior MR exam improved the specificity by 5%, 9%, or 13%, respectively, depending on whether the prior context was from Prior 1 (on average 2 yrs earlier than the index visit), Prior 2 (3 years earlier) or Prior 3 (4 years earlier). Similarly, for the Clinical Only risk refinement framework, prior context from further out in time from the index visit converted more false positives to true negatives. Including additional context improved the specificity by 9% with Prior 2 (on average 2 years earlier than the index visit), and by 15% with Prior 3 (4 years earlier). The inclusion of context from Prior 1 (within 6 months of index visit) only converted 1 false positive to true negative.

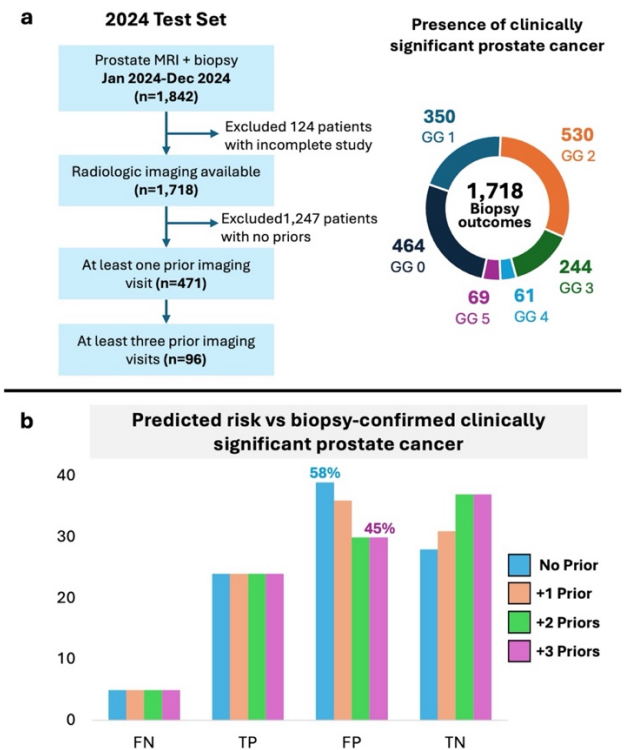

**Figure 5**: Correlation with pathology: evaluation of the 'Imaging Only' risk refinement model performance in predicting the presence of biopsy-confirmed clinically significant prostate cancer (csPCa) on a secondary test set with at least three prior imaging visits. (a) Dataset description. We curated a secondary test set consisting of 1,718 patients who were imaged for known or suspected prostate cancer between January 2024 and December 2024 and subsequently underwent biopsies within 6 months of imaging to test for the presence of csPCa. To demonstrate the effect of integrating prior imaging context on predicting csPCa, we identified a subset of these patients who had at least three prior imaging visits (n=96) between 2015 and 2023. The presence of csPCa is defined as a Gleason Grade (GG) of 2 or higher, where Gleason Grade assesses the aggressiveness of prostate cancer based on histopathological evaluation. (b) Performance comparison. Bar charts compare numbers of false negative (FN), true positive (TP), false positive (FP), and true negative (TN) cases with incorporation of information from 0, 1, 2 and 3 prior visits. Here, the three priors were taken from three consecutive prior imaging visits, with average intervals between index visit and prior visits of 1.8 ± 0.7 years for Prior 1, 3.2 ± 1.1 years for Prior 2, and 4.7 ± 1.3 years for Prior 3.

**DISCUSSION**

In this work, we showed that systematically integrating temporal context into AI-based medical risk prediction significantly reduces false positives without compromising sensitivity. We used a risk-steering approach, assessing initial disease risk based on the most recent imaging examination, and leveraging temporal context gleaned from prior imaging examinations and/or clinical biomarkers to upgrade or downgrade this assessment.

We demonstrated the utility of our context-aware risk prediction approach in predicting the current and future risk of prostate cancer (PCa) in a large cohort of nearly 30,000 men followed for known or suspected PCa with over 39,000 MRI examinations and 68,000 blood tests over the course of nearly a decade. We specifically targeted prostate cancer as a focus for our work, rather than breast or lung cancer, where imaging-based surveillance is typically performed at more regular intervals, in order to demonstrate the robustness of our approach to heterogeneous follow-up intervals. Learning from our large dataset of patients provided our models with useful population-level context which resulted in comparatively high sensitivity. Learning from additional patient-specific priors, even though these priors were far more limited in number and were separated by variable time intervals, provided our models with personalized context to refine risk estimates and improve specificity.

The current healthcare landscape is gradually transitioning from a reactive to an increasingly proactive model, with a rising utilization of longitudinal monitoring approaches, such as active surveillance [6 8], opportunistic screening [19], and more disease-specific proactive screening [20]. Risk prediction models are often optimized for high sensitivity to facilitate early detection and intervention, but this usually comes at the expense of poor specificity, which, in turn, can lead to overdiagnosis, overtreatment [9], and increased health care costs, and can also cause patients to lose faith in active surveillance [11]. There is therefore a need for approaches which can limit false positive findings, particularly in patients with comparatively low *a priori* risk, without compromising the sensitivity of disease detection.

Although longitudinal monitoring can highlight early warning signs, changes over time are not explicitly integrated into medical risk prediction models [21], especially those based on imaging, and are often left for qualitative evaluation by human physicians [22 23]. This makes a rigorous and objective evaluation of risk evolution challenging. The benefit of additional context in improving sensitivity has been demonstrated. For example, recent work in mammography has demonstrated that additional anatomical context from multiple views of the same anatomy during a single visit [24], or from multiple images with complementary contrast [25] improves the sensitivity of automated breast cancer detection. However, the value of integrating additional temporal context, and its particular efficacy in reducing false positives, has not been demonstrated before now.

Our risk refinement approach was loosely modeled on how human clinicians appear to use prior context. When assessing disease risk over time, human physicians typically look for changes from a patient's baseline state of health. They might identify worrisome changes, in which case they will upgrade their risk estimate, or else stable findings, in which case they might downgrade their risk estimate. Consistent with such intuitive expectations, as well as with our starting hypotheses, when integrating risk information over time, our model converted false positives into true negatives, increasing specificity without degrading sensitivity. This general behavior was observed regardless of whether our risk refinement models relied on prior clinical data only, prior imaging data only, or a combination of the two. In all cases, specificity increased further as additional priors were included.

One might ask how such a risk steering approach compares with more generic approaches that simply input all available current and prior information together for risk prediction. We performed such an ablation study, and observed that, compared to models that use all information simultaneously, our risk steering approach proved more effective in converting false positives to true negatives (Supplementary-Figure-S5). Our hypothesis is that, compared to unsteered prediction models that were unable to learn sufficiently robust temporal dynamics, particularly in the setting of limited prior visits with variable gaps, our steered networks had the more focused and more easily learned task of using prior information to adjust a starting risk estimate.

We leveraged representation learning to summarize information regarding underlying risk of disease present in high-dimensional medical data into low-dimensional representations. These learned representations allowed us to 1) predict risk of prostate cancer without relying on any hand-crafted features, explicit organ segmentations or lesion annotation on medical imaging data, and 2) summarize risk trajectories to learn temporal risk changes, without extensive pre-processing approaches such as image registration that are typically used to align medical images acquired across time [26]. This is reflected in saliency maps [27] that highlight regions of interest our model focusses on when making predictions. Even in the absence of explicitly provided lesion information, our model focuses on relevant information including lesions (Supplementary-Figures-S6 and S7) when predicting risk using imaging examinations from current and prior visits.

Our primary goal was to show how integrating prior context can improve specificity in medical risk prediction. For our selected use case of PCa risk prediction, we used radiological assessments (i.e., PI-RADS ratings) as our ground truth, with PI-RADS values of 4 or 5 indicating high risk (positive). Though PI-RADS ratings of 4 or 5 generally indicate a high likelihood of clinically significant PCa, Gleason scores derived from prostate biopsies are usually considered more reliable ground truth assessments. However, Gleason scores were available only for a comparatively small subset of our patient cohort, whereas radiologist-assigned PI-RADS scores are available for all patients in our cohort, providing a more robust training set. Previously, we showed that our PI-RADS-trained representation learner model was able to separate low-risk from high-risk representations, achieving good correlation with associated biopsy outcomes [28]. Benign or negative biopsies were mostly clustered in the low-risk region of the learned representation space, and high-grade cancers were clustered in the high-risk region. Although trained on PI-RADS-based ground truth, our model's risk assessment correlated well with the presence of clinically significant prostate cancer as established by biopsy in a subset of patients for whom this information was available.

Although all our training and evaluation data was collected from a single health care enterprise, it is a comparatively large and distributed enterprise, with data reflecting real and evolving clinical practice on the ground. Our large-scale prostate monitoring data collected from more than 28,000 men consisted of patients of varying ages and overall states of health, observed over a long study period. The imaging data were collected from many disparate models of MRI scanners (Siemens Aera, Prisma, Prisma fit, Skyra, Vida, Tim Trio, Verio and Verio DOT) across 56 imaging centers, over a sufficiently long time period to encompass significant changes in technology and practice.

We demonstrated that our risk refinement approach is generalizable to medical risk prediction models utilizing different kinds of longitudinal data (imaging and/or clinical biomarkers). It would be of interest in future to evaluate the performance of this risk refinement framework in longitudinal datasets from external sites, and also to evaluate its performance in a prospective setting. However, this is a rather challenging proposition at present, as longitudinal datasets that track patient health outcomes over long periods of time are difficult to come by and nontrivial to curate, and as prospective assessments of risk prediction in slowly evolving diseases like prostate cancer require lengthy evaluation periods. We hope and expect that suitable longitudinal datasets will become increasingly available as time goes on, as interest in proactive health continues to grow.

In this work, patient-specific representations used for risk-assessment were derived from a single radiological imaging modality (MRI) and a collection of clinical variables that are known to be of interest for PCa risk stratification [13 15]. Future expansions of our risk refinement framework could include extraction of useful contextual information from other imaging and sensing modalities as well as other previously acquired clinical data (e.g., patient demographics, digital rectal examination results, reports of surgical interventions, radiology or histopathology reports, physician notes, proteomics, lesion genomics, etc.). Static baseline risk-modifying data such as polygenic risk scores [29] should also be relatively straightforward to incorporate into our risk refinement framework. As evidenced by our results for combined clinical and imaging priors, it is expected that, with suitable approaches such as the successive refinement used here, larger quantities of diverse contextual information should only improve risk assessment, up to certain limits given by the degree of mutual information. We also expect that related risk refinement approaches should add value for other dynamic disease processes such as breast cancer, lung cancer, pancreatic cancer, Alzheimer's disease and other neurodegenerative diseases, etc. Finally, the use of suitable contextual information may have the additional advantage of increasing the accessibility of longitudinal monitoring. Preliminary evaluations suggest that context may enable accurate risk prediction from current data of limited quality and/or quantity. Recent advances in accessible MRI, for example, raise the prospect of using MRI for more broad-based longitudinal monitoring with low-quality MR images from cheap but accessible scanners, backed up by higher-quality images and diverse clinical data as context.

In the long run, reduction of false positive rates to levels commensurate with the prevalence of common diseases in the general population would enable effective population-level longitudinal health monitoring. This would shift the current episodic paradigm of healthcare still further towards a proactive and continuous model. In the meantime, we have demonstrated an AI-based risk refinement strategy which successfully incorporates diverse temporal context to drive down false positive rates in an existing setting of longitudinal monitoring for at-risk individuals. These results suggest that the statistics of health monitoring are not inherently static. Instead, the combination of modern machine learning with modern medical sensing technology affords the opportunity to connect disparate episodes of care, improving outcome prediction progressively as more information is gleaned about an individual over time.

## METHODS

### Data curation

For our AI model training and evaluation, we retrospectively collected imaging and clinical data (with approval from and oversight by our Institutional Review Board) from patients who underwent prostate MR imaging for known or suspected prostate cancer between 2015 and 2023 on any MR scanner in our health enterprise (Supplementary Figure S1). The MR scanners in question, though all Siemens equipment, represented a range of scanner types (Siemens Aera, Prisma, Prisma fit, Skyra, Vida, Trio Tim, Verio and Verio DOT) and software levels, and were spread across 56 imaging centers in the New York metropolitan area. Imaging data from a given visit were included if the MR examination was complete, if bi-parametric MR images (consisting of T2-weighted images (T2WI) and Diffusion-weighted images (DWI)) were available, and if a PI-RADS score had been provided by radiologists. The inclusion criteria were satisfied by 28,342 men (mean age = 65.7±8.4 years) with 39,013 MR examinations and associated radiology reports. We also retrospectively collected 68,931 PSA tests from a subset of these patients (n=13,764) who also had longitudinal monitoring with serum PSA values during the study period.

In our cohort, radiologic imaging for 21,550 patients included data from only one point in time with no follow-up imaging. 6,792 patients (17,463 MR exams) had at least one follow up visit, with an average time interval of 1.8±1.1 years between visits (min=0 years, max=8 years). Since our risk prediction model assumes that risk monotonically increases with time in the absence of any intervention (see Risk Estimation model), we excluded any patients whose PI-RADS score was lowered from a high-risk category (PI-RADS 4 or 5) to a low-risk category (PI-RADS 1-3). The remaining 6,381 patients with longitudinal imaging data and associated clinical data were randomly split into training (n=5,167 men; 13,121 MR exams), validation (n=579 men; 1,482 MR exams), and test (n=635 men; 1,612 MR exams) sets. We also associated current and prior clinical data (PSA, age, and prostate volume) with each of these imaging-based index visits (Supplementary Figure S1b). Although the prostate volumes used here were derived from length, width, and height measurements reported in the radiology reports, these may also be available from text reports and/or ultrasound exams, and hence, were included with clinical data.

**Ground truth**

For each visit, we used the associated maximum PI-RADS score for all examined images as a surrogate for the patient-level risk of prostate cancer. A high risk of prostate cancer (positive label) was defined as a max PI-RADS of 4 or 5. (See Discussion for comments on the rationale for this choice and the correlation of these labels with Gleason scores.)

**Temporal augmentation**

For each patient with imaging data at multiple timepoints, we randomly selected data from a timepoint of interest as the index visit. All medical data (imaging and clinical) prior to this index visit were considered prior information, and radiologic assessments corresponding to subsequent MR exams were considered future diagnosis. By temporally shifting this selection window, we performed temporal augmentation on our longitudinal imaging data to further increase the size of training and evaluation datasets available to us. This training set augmentation over time resulted in 13,121 MRI exams with ground truth assessments available for current risk assessment, and 5,361 MRI exams assessments for 5-year risk (Figure 2). As a result of the augmentation strategy, some data from patients with multiple visits was used more than once with different temporal shifts. However, training, validation and test sets remained rigorously separate at the patient level.

**AI Model**

Our risk refinement framework (Figure 1) uses clinical and/or radiologic imaging data from the most recent patient visit to predict high risk of prostate cancer (PCa) at the time of the index visit, and within 5-years of the index visit, and leverages patient-specific information from prior visits, whenever available, to refine this initial assessment. This framework consists of a representation learner (RL) model that transforms clinical data or high-dimensional imaging data into subject-specific latent representations, a risk estimation (RE) model that predicts PCa risk based on the most recent representation, and temporal learner (TL) and risk refinement (RR) models that aggregate temporal history into a change signal to upgrade or downgrade the initial risk assessment. Each of these models is described in detail below:

*Learning risk representations with a Representation Learner (RL)*

Two distinct RL modules were used, one for imaging data and another for clinical data.

The image RL model (Supplementary Figure S8a-c) uses T2WI and DWI prostate volumes as inputs to generate a 256-dimensional summary representation. This RL model consists of two 3D convolutional neural network (CNN) encoders that independently transform T2WI and DWI data, respectively, to generate contrast-specific representations, and a transformer encoder that aggregates them into a subject-specific representation. Each MRI contrast volume (T2WI and DWI) is processed independently without any explicit registration.

We previously showed that the RL model, when pretrained with PI-RADS-guided contrastive learning, can learn a mapping from prostate MR images to a low-dimensional representation space that is explanatory of a radiologist's assessment of these images [28]. We use the same approach in this work to initialize the weights of our RL model and generate subject-specific representations. The RL model was trained with PI-RADS guided contrastive loss such that the similarity between representations of a pair of prostate MRI volumes with same PI-RADS assessments were maximized. In this learned representation space, images with similar radiological assessments are pulled together: i.e., representations from no- or low-risk assessments (PI-RADS 1 or 2) are pulled towards one another and pushed away from high-risk assessments (PI-RADS 4 or 5).

For the clinical risk refinement model, we replaced the image representation learner with a smaller clinical encoder (Supplementary Figure S8e) that transformed clinical data into a 32-dimensional summary representation.

*Risk estimation (RE) with an additive hazard model*

Given a subject specific representation $z$ emerging from the RL module, the RE model predicts the risk of PCa associated with z at the time of the index visit and within 5-years. We use an additive hazard model, in line with recent risk prediction works in the literature [24 25], and define the probability of having a high-risk of PCa within time k (years) in the future conditioned on $z$ as follows:

$$P(t = k|z) = \sigma(R_k) = \sigma\left(B_0(z) + \sum_{i=1}^{k} H_i(z)\right)$$

This additive hazard model first estimates a baseline risk $B_0(z)$ followed by marginal increases in hazard $H_i(z)$ over time $t_i$ (5-years in the future in our case). The baseline risk and marginal increases in risk are predicted by a series of dense layers. Here, the operation $\sigma$ is a sigmoid function that converts predicted risk scores to probabilities. Instead of independently predicting risks at current and future timepoints, this cumulative formulation ensures that the future risk of disease is always greater than or equal to the current risk of disease.

*Risk refinement (RR) with a temporal learner (TL)*

For a sequence of longitudinal data from current and prior visits of a subject, we use the trained RL model to generate a sequence of risk representations (Figure 1). The TL model is a transformer encoder (Supplementary Figure S8d) that ingests an arbitrary number of subject-specific temporal representations along with their associated times, to learn an auxiliary change signal 'g' to steer risk via the RR model. We use the time intervals between the index visit and its priors as positional embeddings to encode time and provide the model with temporal context for understanding the given sequence of representations. For each representation, we discretized the associated time information to yearly intervals for the imaging data (max 10 years) and quarterly intervals for the blood tests (max 40 quarters). Motivated by the five-point scoring system in PI-RADS to describe the likelihood of clinically significant PCa, and the five-point scale in PRECISE to describe the likelihood of radiological changes in serial imaging during active surveillance [22 23], we set the dimensionality of the change signal g to 5.

The risk refinement model mimics the architecture of the risk prediction model described previously but is conditioned on the personalized change signal g to predict changes in risk at time t given the change signal g learned from patient-specific priors. Incorporating temporal information, the final probability $P_f$ of high-risk PCa in k years is given by

$$P_f(t = k|z, g) = \sigma(R_k + \tilde{R}_k)$$

Here, $R_k$ and $\tilde{R}_k$ refers to the predictions from the RE model and the RR model, respectively.

**Data processing**

The MR image acquisition parameters for T2WI and DWI are described in Supplementary Table S6. The DWI data consisted of an apparent diffusion coefficient (ADC) map and a high b-value image (b=1500 s/mm$^2$). Each MRI volume was normalized to have signal intensities in the range [0,1] and resampled to matrix size 256x256x24, followed by a 3-dimensional crop to retain the central 128x128x16 region containing the prostate. For training the AI model, data augmentations consisted of random left-right flips, in-plane rotations in the range $[-10°, 10°]$, in-plane elastic deformations, and jittered central crop by varying the resampling matrix sizes in the range [224, 284]. Clinical data included serum PSA values, age, and prostate volume. As the distribution of raw PSA values is heavily skewed towards low values, we transformed serum PSA values logarithmically before the analysis to normalize the data distribution [30], and to emphasize relative changes over absolute changes in PSA values over time. Prostate volumes were obtained using the length, width, and height measurements reported in the associated radiology reports. The clinical data was then z-score normalized using mean and standard deviation values calculated from the training population.

**End-to-end risk assessment**

To demonstrate the generalizability of our temporal risk refinement approach to different types of data, we trained three different models for PCa risk prediction: a) 'Clinical Only' refinement model that used only clinical priors, b) an 'Imaging Only' risk refinement model that used only imaging priors, and c) an 'Imaging + Clinical' multi-modal risk refinement model that leveraged both imaging and clinical priors.

*Risk assessment with clinical data*

The clinical RL, RP, TL, and RR models were trained in an end-to-end supervised fashion, using pairs of randomly sampled timepoints and the clinical data corresponding to them. The time intervals between recent PSA and prior PSA tests in months were converted to quarterly intervals, with the upper bound on time set to 10 years (40 quarters). We first trained the model to predict current risk, followed by the 5-year risk using data for which these assessments were available.

*Risk assessment with imaging data*

For image-based risk refinement, we used a multi-stage training approach to ensure relevant risk features were extracted from high-dimensional medical images. The RL model was pretrained with the PI-RADS guided contrastive strategy described previously in Umapathy et al [28]. We used MR exams and the associated labels from patients with no longitudinal follow-up (n=21,550) to pretrain the model to learn a PI-RADS-guided representation space such that representations of MRI exams with similar PI-RADS assessments are pulled closer to one another. We then used the temporally augmented longitudinal prostate MR images (n=13,121) for fully supervised training of the RL and RP models together to predict the current and 5-year risk of PCa using prostate MR images from the index visit. The entire risk refinement framework, including the temporal learner and the RR models, was then trained end-to-end with supervision. For each training iteration, we used pairs of temporal data that included an index visit and a randomly selected prior visit to learn steering of initial risk based on variable time intervals between current and prior visits.

*Multi-modal risk assessment*

For leveraging information from both imaging and clinical data collected from prior visits, we used a successive refinement approach. As before, the representation from the most recent imaging data was used for initial risk estimation. We then used the trained temporal learner and risk refinement models from the imaging-only priors and clinical-only priors to learn changes in risk from prior images and prior clinical data, respectively. The initial risk assessment was updated first using changes in risk based on imaging priors, followed by changes in risk based on clinical priors.

**Model training details**

Each contrast-specific encoder (for T2WI and DWI) in the RL model was independently pretrained with the PI-RADS guided contrastive loss, followed by end-to-end pretraining of the RL model using the same loss function as described in Umapathy et al. [28]. During pretraining, the models were trained for 50 epochs, with an initial learning rate of 0.001, stochastic gradient descent optimizer, and the temperature tau for contrastive loss was set to 0.07. The entire risk prediction and refinement framework was trained fully supervised end-to-end using weighted categorical cross entropy (WCCE) loss as shown below.

$$L(r,p) = -\sum_n w_n(1-r)\log(1-p) + w_p r \log(p)$$

Here, $n$ is the number of samples, $r$ and $p$ correspond to the ground truth label and predicted risk probabilities, and the weights $w_n$ and $w_p$ correspond the computed weights of the negative and the positive labels such that the weights were inversely proportional to the label frequencies in the training dataset. Since we had 13,171 labels for current risk, and only 5,361 labels for 5-year risk, we used a masked variant of WCCE to only compute loss for data points where the labels were available. This was done using a mask variable $m$ associated with each label and timepoint (current and future) where the mask $m$ was set to 1 or 0 depending on whether an image-based ground truth assessment was available for that time. For the end-to-end supervised finetuning, the models were trained for 100 epochs with the optimizer set to AdamW, with an initial learning rate of $10^{-5}$ and weight decay of $10^{-4}$. The learning rate was halved whenever the validation loss plateaued for 5 consecutive epochs, with checkpoints saved for the lowest validation loss. We also incorporated early stopping to avoid the model from overfitting. All models were trained on a single NVIDIA A100 80GB GPU and all implementations were coded in Python using Keras with Tensorflow backend (https://www.tensorflow.org/).

**Evaluation and Statistical analysis**

The model performances were evaluated on distinct test sets (Supplementary Figure S1c) based on the type of prior data (Clinical Only, Imaging Only, Imaging + Clinical). The performance of the models with and without prior context was evaluated using receiver operating characteristic curves. We calculated the 95% confidence intervals (CI) for AUC using DeLong's approach [31 32]. The operating points (i.e., thresholds for calling a given case positive) were selected for target sensitivity levels based on model performances on the validation sets. The target sensitivity levels were set at 85% for the Clinical Only model, and 90% for the imaging-based models and metrics such as sensitivity and specificity were computed. McNemar tests were used to compare significant differences in sensitivity and specificity of the different models. We used Cohen's kappa score for measuring agreement between the models in identifying true positives with and without prior context. The value of alpha was set to 0.05 for the statistical tests.

**ACKNOWLEDGEMENTS**

This study was performed under the rubric of the Center for Advanced Imaging Innovation and Research (CAI²R, www.cai2r.net), an NIBIB National Center for Biomedical Imaging and Bioengineering (NIH P41 EB017183).

**Data Availability:** The data used in this work are not available publicly yet. We are currently exploring appropriate deidentification strategies for public sharing, given that large longitudinal datasets such as the ones used in this work are at present comparatively rare and valuable.

**Conflicts of Interests:** The authors report no conflicts of interest relevant to this manuscript.

# Supplementary Material for

# Context-aware deep learning using individualized prior information reduces false positives in disease risk prediction and longitudinal health assessment

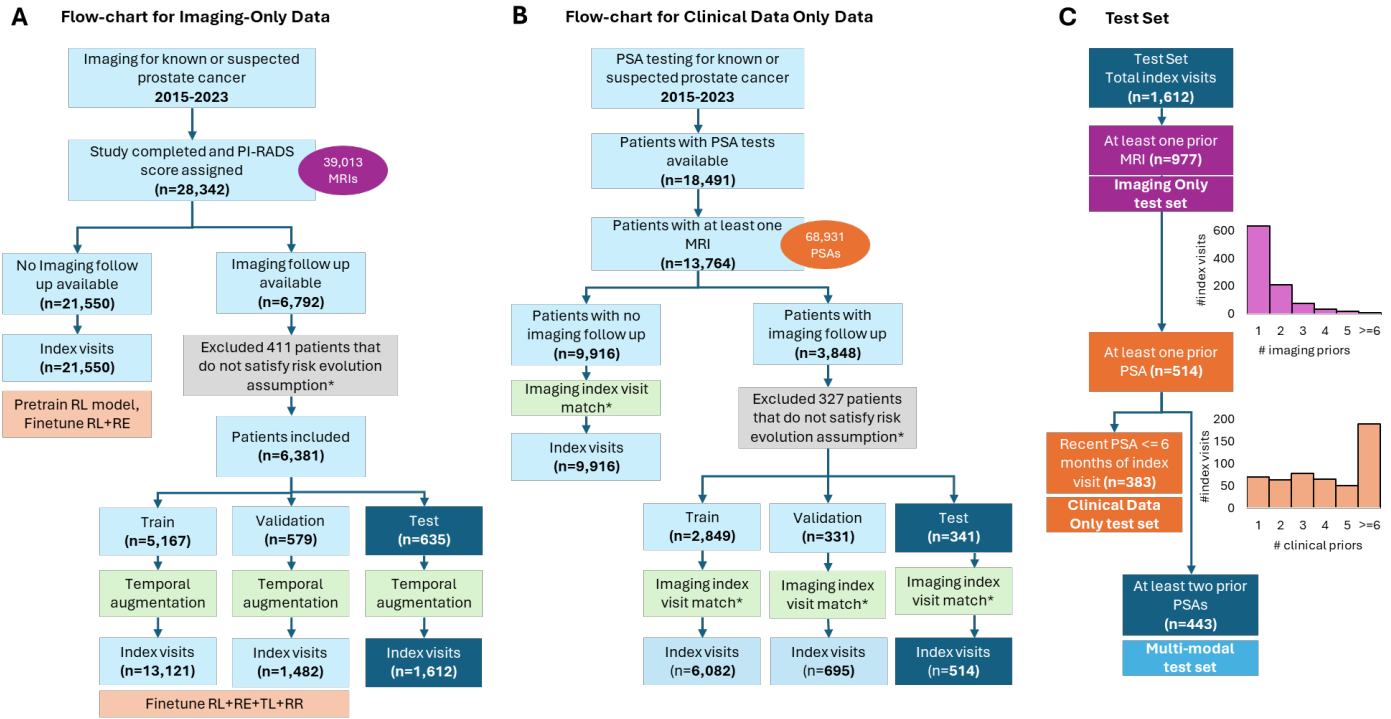

**Figure S1:** Flow-chart for curation of prostate cancer (PCa) longitudinal dataset used for temporal risk refinement. (A) Curation of radiologic imaging data for the 'Imaging Only' risk refinement model. All patients who were imaged for known or suspected prostate cancer between 2015 and 2023, and had complete bi-parametric MRI examination with a PI-RADS score assigned were included. The patients were split into two groups – those with longitudinal imaging follow-up and those without. Patients with MRI examination from only one point in time were used to train the image representation learner (RL) and risk estimation (RE) model to predict the risk of PCa using population context. Our risk prediction model assumes that risk of disease increases monotonically over time in the absence of any intervention. As such, 411 patients with longitudinal follow up that did not satisfy this risk evolution assumption were excluded. The temporal data from the remaining 6,381 patients were used to learn risk refinement based on personalized context. These patients were randomly assigned into training, validation, and test splits. We used temporal augmentation to identify index visits with priors and future assessments. (B) Curation of clinical data for the subset of patients in (A) that also had prior blood tests for prostate-specific antigen (PSA) available. (C) Flow-chart for the temporally augmented test set. The index visits for the test sets (n=1,612, 635 patients) consisted of 977 imaging visits with at least one imaging prior, 342 imaging visits with at least two imaging priors, 135 visits with at least three imaging priors. The test set also consisted of 514 visits with at least one clinical prior, and 443 visits with both imaging and clinical priors. For the Clinical Only test set, we only selected visits where the recent PSA test was performed within 6 months of the imaging index visit.

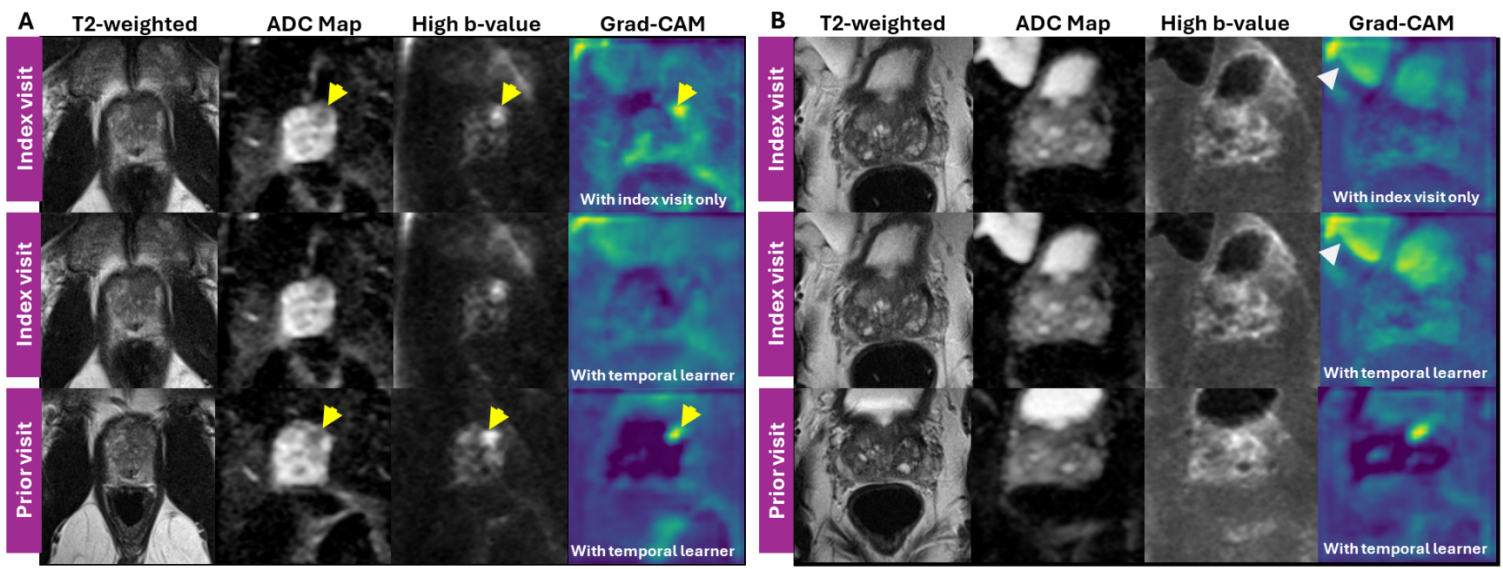

**Figure S2:** Gradient-weighted Class Activation Mapping (Grad-CAM) saliency maps for a subject mis-classified (false negative) by the 'Imaging Only' model when incorporating prior imaging context. T2-weighted image, Apparent Diffusion Coefficient (ADC) map, High-b-value image, and Grad-CAM saliency maps for a 62y old patient (age at index visit) assessed as high risk for clinically significant prostate cancer (PI-RADS 4) are shown here. (A) This patient had an apex transition zone lesion (highlighted by the yellow triangle in ADC map) that was upgraded from equivocal risk (PI-RADS 3) from a prior visit three years ago (age at prior visit = 59y). When using MRI from the index visit, the Imaging Only model assessed risk as positive, and Grad-CAM saliency map showed focus on the transition zone lesion (top row). When adding context from a prior MRI from 3 years earlier (bottom row), the risk refinement model mis-classified this patient as negative. Grad-CAM saliency maps showed that the temporal learner, that was trained to leverage temporal context by looking at changes between current and prior MRI exams, was not looking at the lesion in the index MRI (middle row). (B) On further investigation, we found that the temporal learner was focusing on another significant change between the two MRI exams instead – a penile prosthesis reservoir implant which was not present previously. Although in this case the model focused on an irrelevant change resulting in misclassification, this further highlights that risk prediction models need relevant context (in this case – presence of a surgical implant) to make better decisions.

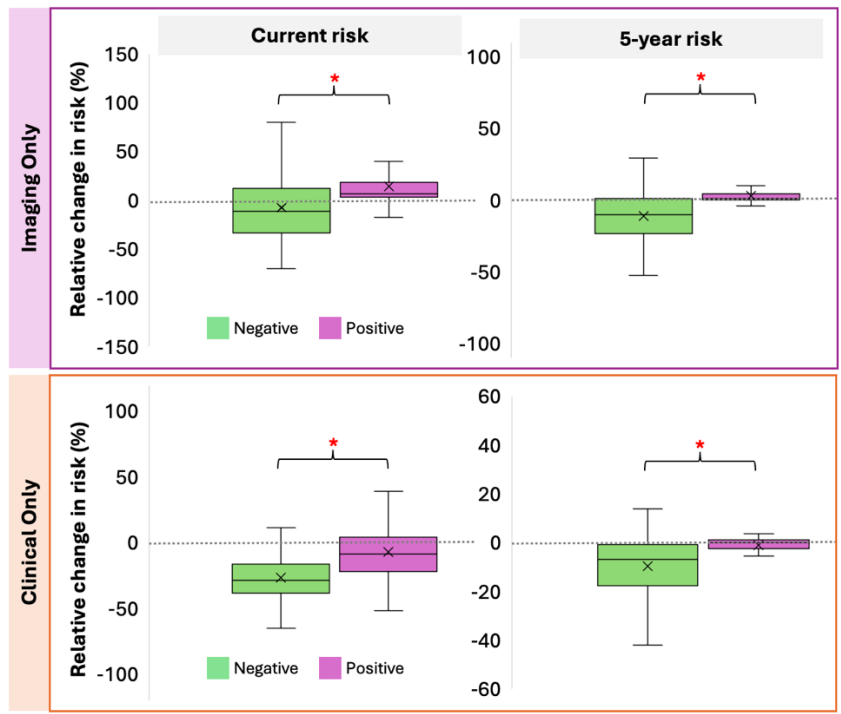

**Figure S3**: Visualization of differential risk steering with prior context. The box and whiskers plots show changes in risk assessments for negative and positive cases for the 'Imaging Only' risk refinement model (top) and the 'Clinical Only' risk refinement model (bottom). On incorporating prior context, risk refinement frameworks differentially steered risk estimates based on the index visit for the negative and positive cases. On average, the Imaging Only risk refinement model significantly downgraded the current risk estimates based on the index visit for negative cases (relative change in risk probabilities with respect to index visit = -8%) and upgraded the current risk estimates from the index visit for positive cases (mean=+14%). Here, the red asterisk denotes a significant difference ($p<0.001$, unpaired two-sided Student's t-test) in mean relative change in risks between the positive and negative cases. Relative change in risk was calculated as 100 * (initial risk – refined risk)/initial risk. Here, initial risk is the risk predicted from the index visit only, and the refined risk is the risk predicted when integrating context from one prior visit. The dotted gray line denotes no relative change in risk. For the 5-year risk, the model downgraded initial risk estimates on average for negative cases (-11%) and upgraded initial risk estimates for positive cases by (+3%). This difference in risk refinement between the positive and negative cases was significant for both current risk and the 5-year risk of PCa ($p<0.001$, unpaired two-tailed Student's t-test). We also observed similar significant differences between negative and positive cases for the Clinical Data Only risk refinement model.

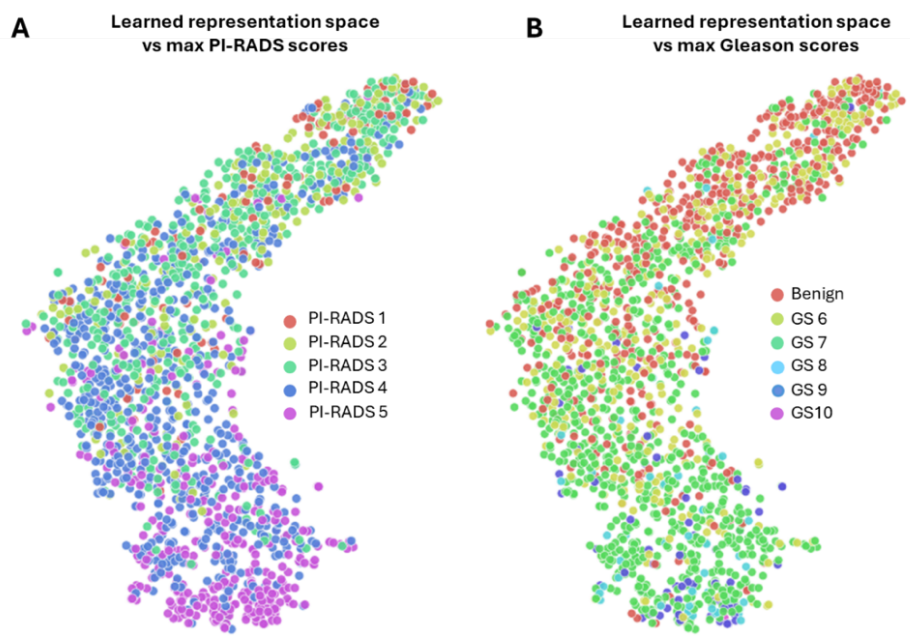

**Figure S4:** Visualization of learned risk representation space. The scatterplot presents t-Stochastic Neighborhood Embedding (t-SNE) visualization of representations generated by the image representation learner (RL) model for bi-parametric prostate MR images from a secondary test set. We curated a secondary test set consisting of 1,718 patients who were imaged for known or suspected prostate cancer between January 2024 and December 2024, and subsequently underwent biopsies within 6 months of imaging to test for the presence of clinically significant prostate cancer. The RL model was trained with PI-RADS guided contrastive learning to summarize images into latent representations such that they are indicative of the underlying risk of prostate cancer. The generated representations are color-coded by the radiologists' risk assessments (maximum PI-RADS per MRI) on the left (A), and the histopathological outcomes (maximum Gleason scores per MRI) on the right (B). The learned representation space separates no/low-risk representations (PI-RADS 1-2) from high-risk representations (PI-RADS 4-5). These also visually correlate with the biopsy outcomes. We also note that the RL model moved some PI-RADS 4 assessments to the low-risk region (blue dots) in panel A, and these corresponded with benign biopsies in panel B (red dots). This shows that the model learns a risk space that correlates well with biopsy outcomes as well.

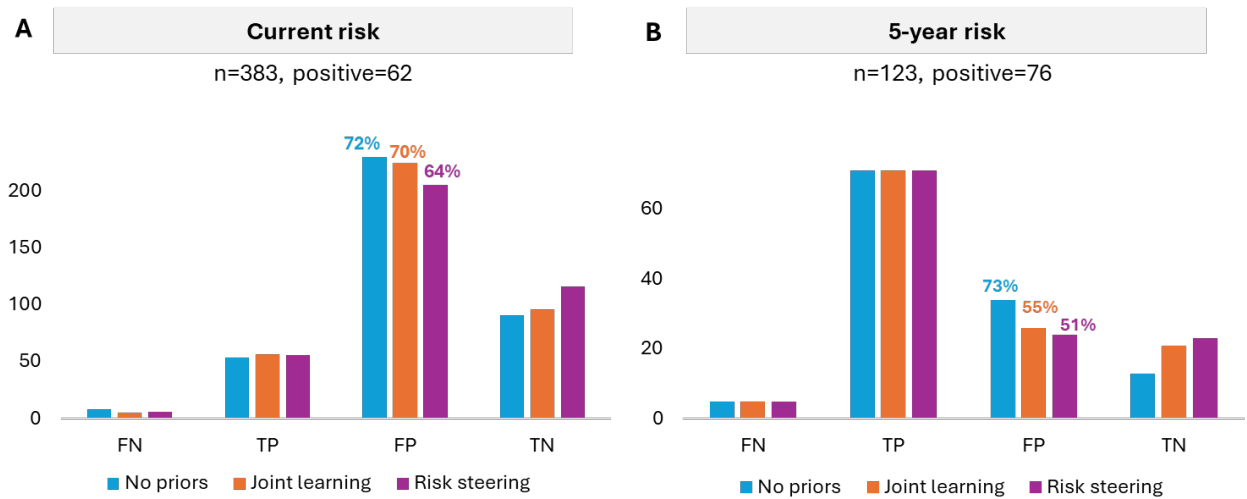

**Figure S5:** Joint learning from current and prior data vs risk steering with prior data – an ablation study on 'Clinical Only' risk refinement model. The bar-plots compare the performance of three different risk prediction models that use clinical data from an index visit and a prior visit to predict risk of prostate cancer at the time of the index visit (current risk), and within 5-years of the index visit (5-year risk). Here, the prior visit corresponds to the first visit (oldest prior). The false positive rates associated with each model are noted (and color-coded) above the false positive bars. The first model uses representations from clinical data (serum prostate-specific antigen values, age, prostate volume) from the index visit. The joint learning model uses a clinical representation learner followed by a temporal learner (same architecture as the proposed work) to jointly learn to predict risk of PCa. The risk steering model (proposed approach) uses the temporal representations generated by the clinical representation learner to steer risk estimates made from representations of the index visit. We observe that utilizing temporal context (joint learning and risk steering) reduces false positive rates and increases specificity. However, risk steering results in increased specificity compared to just jointly learning from all temporal data.

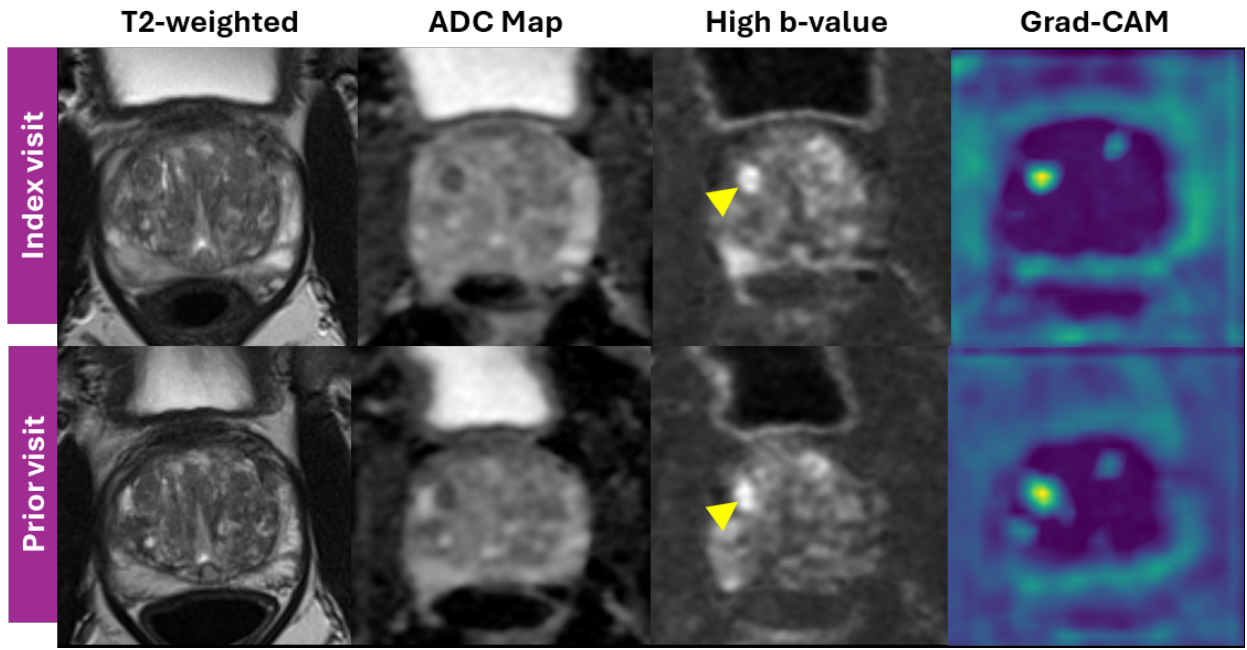

**Figure S6:** Gradient-weighted Class Activation Mapping (Grad-CAM) saliency maps for a subject that was downgraded from a false positive to true negative on integrating prior imaging context. T2-weighted image, Apparent Diffusion Coefficient (ADC) map, High-b-value image, and Grad-CAM saliency maps of the prostate for a 61y old patient (age at index visit) with no suspicious findings (PI-RADS 1) on the MRI examination are shown here. The MRI examination was mis-classified as positive (false positive) by the 'Imaging Only' model when using information from the index visit. The Grad-CAM saliency map (top row) for the index visit showed that the model was focusing on the hyperintense region in the transition zone (highlighted by the yellow triangle on the high b-value image). On integrating MRI (bottom-row) from a year ago (age at prior visit = 60y), our risk refinement model downgraded the current risk of prostate cancer (relative change in risk = -47%), resulting in the case being classified correctly as negative for clinically significant prostate cancer.

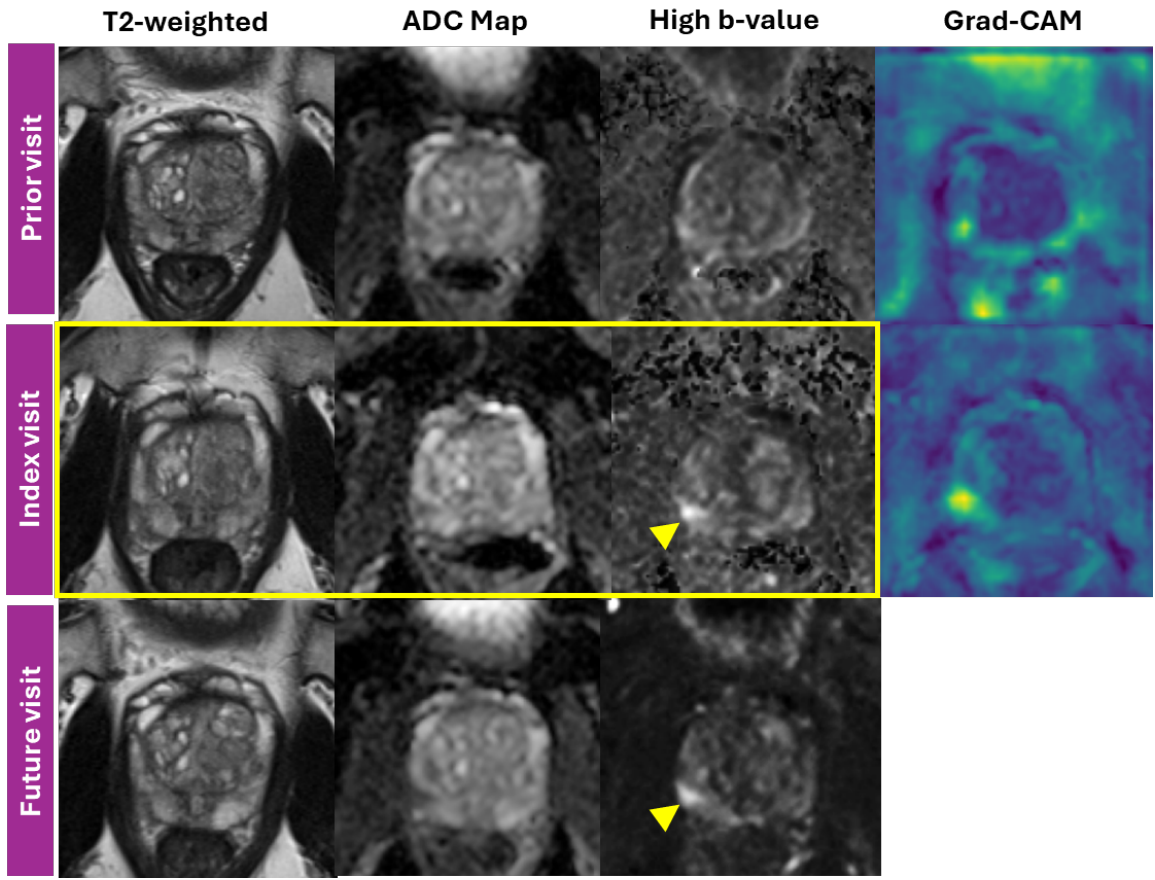

**Figure S7:** Gradient-weighted Class Activation Mapping (Grad-CAM) saliency maps for a subject with an equivocal assessment (PI-RADS 3) at the index visit and clinically significant prostate cancer in the follow-up visit. T2-weighted image, Apparent Diffusion Coefficient map, High-b-value image, and Grad-CAM saliency maps of the prostate for a 77y old patient (age at index visit) assessed as equivocal risk (PI-RADS 3) are shown here. The MRI examination at the index visit (middle row) had a right posterolateral mid gland peripheral zone lesion (highlighted by the yellow triangle in the high b-value image) that was stable with respect to prior MRI from a year ago (top row). The bottom row shows the MRI examination from 3 years later (age at future visit = 80y). Although the lesion was stable in size, it was deemed more 'conspicuous' compared to previous examinations and upgraded to PI-RADS 4. A follow-up biopsy revealed that the lesion was Gleason 3+4 (clinically significant prostate cancer). On integration MRI from a prior visit (top row) one year ago with the index visit, our risk refinement model upgraded the 5-year risk of PCa (relative change in risk = +30%). Grad-CAM saliency maps revealed that the temporal learner was focusing on the lesion of interest in both the index and prior MRI exams.

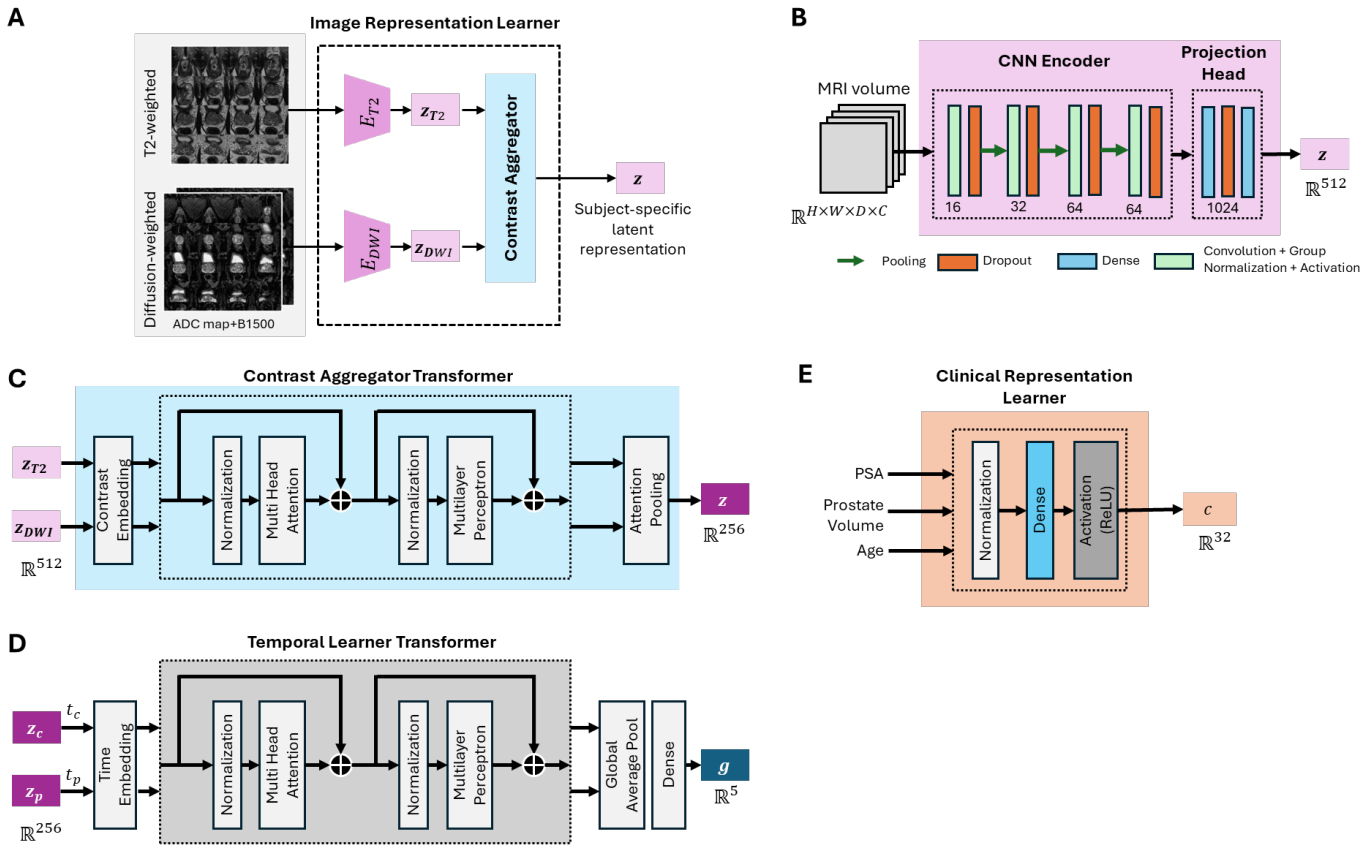

**Figure S8:** Architectures of the deep learning models that constitute our risk refinement framework. (A) An image representation learner (RL) summarizes information from bi-parametric prostate MR images into latent representations that are indicative of underlying risk of disease. We use a PI-RADS guided contrastive learning approach to pretrain our representation learner. The RL model consists of (B) CNN encoders ($E_{T2}$ and $E_{DWI}$) to independently transform T2-weighted and multi-channel Diffusion-weighted images into latent representations $z_{T2}$ and $z_{DWI}$. A contrast-aggregator transformer (C) learns to combine these contrast-specific representations into a subject-specific representation $z$. (D) A temporal learner transformer takes a sequence of subject-specific temporal representations $z_c$ and $z_p$ from current and prior visits, along with their times $t_c$ and $t_p$, to generate an auxiliary signal g that is used to steer risk estimation from current risk representation $z_c$. For the Clinical Data Only model, we replace the image RL with a clinical representation learner shown in (E). The clinical RL transforms clinical data (serum prostate-specific antigen, age, and prostate volume) from a given visit into a latent representation $c$. Here, the clinical temporal learner has the same architecture as the image-based temporal learner but with a lower dimensionality (d=32 vs d=256).

**Table S1:** Data characteristics of our prostate cancer longitudinal monitoring dataset.

|  | All | Imaging-follow up (Train) | Imaging-follow up (Validation) | Imaging-follow up (Test) |
|---|---|---|---|---|
| # Patients | 28, 342 | 5,167 | 579 | 635 |
| Age (y) | 66.3 ± 8.4 | 66.3 ± 7.8 | 66.6 ± 7.7 | 67.9 ± 7.7 |
| # imaging visits (MRI volumes) | 39,013 | 13,121 | 1,482 | 1,612 |
| Risk stratification (study level) | | | | |
| # PI-RADS 1 (%) | 11,377 (29.2) | 3777 (28.8) | 452 (30.5) | 508 (31.5) |
| # PI-RADS 2 (%) | 10,197 (26.1) | 4215 (32.1) | 440 (29.7) | 475 (29.5) |
| # PI-RADS 3 (%) | 8,625 (22.1) | 3213 (24.5) | 359 (24.2) | 383 (23.6) |
| # PI-RADS 4 (%) | 5,567 (14.3) | 1311 (9.9) | 173 (11.7) | 168 (10.4) |
| # PI-RADS 5 (%) | 3,247 (8.3) | 605 (4.6) | 58 (3.9) | 78 (4.8) |
| Risk transitions (patient level) | | | | |
| # No change (low risk) (%) | - | 4,151 (80.3) | 459 (79.3) | 505 (79.5) |
| # No change (high risk) (%) | - | 633 (12.2) | 71 (12.3) | 84 (13.2) |
| # Change (low to high risk) (%) | - | 383 (7.4) | 49 (8.4) | 46 (7.2) |
| Clinical data | | | | |
| Prostate volume (cc) | 65.8 ± 42.2 | 69.9 ± 41.3 | 69.3 ± 41.9 | 71.5 ± 51.8 |
| # Patients with PSA | 13, 764 | 3,112 | 357 | 379 |
| # PSA tests | 68, 931 | 21,813 | 2,567 | 2,653 |
| # imaging visits with prior PSAs | 15,422 | 6,082 | 695 | 514 |
| PSA (mean ± std) ng/mL | 7.3 ± 42.9 | 6.7 ± 22.9 | 7.5 ± 16.2 | 6.7 ± 7.6 |
| PSA (min, max) ng/mL | (0.0, 4042.0) | (0.006, 2604.0) | (0.006, 506.7) | (0.006, 232.6) |

MRI = magnetic resonance imaging; PI-RADS = Prostate Imaging Reporting and Data System; high-risk = PI-RADS 4 and 5; cc = volume of prostate in cubic centimeters; PSA = prostate specific antigen; ng/mL = PSA measurements in blood in nanograms per milliliter

Table S2: Predicting high risk of prostate cancer at the time of the index visit: Performance comparison of risk refinement models. For each index visit during which imaging was performed, we identify test cases with at least three prior MRI exams for the image-based risk refinement model and with a recent PSA test (within 6 months of the index visit) and three prior PSA tests for the clinical risk refinement model. For Clinical Only, the three prior visits (Priors 1, 2, and 3) were selected such that Prior 1 was the most recent visit (average time interval from index visit = 7 ± 6 months), Prior 3 was the initial visit (47 ± 25 months), and Prior 2 was the midpoint (25 ± 14 months). For Imaging Only, the three prior exams were taken from three consecutive prior imaging visits, with average intervals between index visit and prior visit of 2.1 ± 1.2 years for Prior 1, 3.3 ± 1.3 years for Prior 2, and 4.6 ± 1.4 years for Prior 3.

|  | False negatives | True positives | False positives | True negatives | Sensitivity | Specificity | AUC |
|---|---|---|---|---|---|---|---|
| Clinical data-based starting risk refined with clinical priors ("Clinical Only," n=279, positive =44) | | | | | | | |
| No prior | 3 | 41 | 176 | 59 | 0.93 | 0.25 | 0.81 |
| +1 Prior | 3 | 41 | 175 | 60 | 0.93 | 0.25 | 0.82 |
| +2 Priors | 3 | 41 | 163 | 72 | 0.93 | 0.31 | 0.83 |
| +3 Priors | 3 | 41 | 145 | 90 | 0.93 | 0.38 | 0.84 |
| Image-based starting risk refined with imaging priors only ("Imaging Only," n=135, positive=14) | | | | | | | |
| No prior | 1 | 13 | 61 | 60 | 0.93 | 0.49 | 0.87 |
| +1 Prior | 1 | 13 | 56 | 65 | 0.93 | 0.54 | 0.88 |
| +2 Priors | 1 | 13 | 42 | 79 | 0.93 | 0.65 | 0.89 |
| +3 Priors | 2 | 12 | 40 | 81 | 0.86 | 0.67 | 0.88 |
| Image-based starting risk refined with imaging and clinical priors (whenever available) ("Imaging + Clinical," n=135, positive=14) | | | | | | | |
| No prior | 1 | 13 | 61 | 60 | 0.93 | 0.49 | 0.87 |
| +1 Prior | 1 | 13 | 44 | 77 | 0.93 | 0.64 | 0.88 |
| +2 Priors | 1 | 13 | 34 | 87 | 0.93 | 0.72 | 0.89 |
| +3 Priors | 2 | 12 | 29 | 92 | 0.86 | 0.76 | 0.88 |

AUC = area-under-receiver-operating-characteristic curves; No prior = risk prediction with index visits only; +1 Prior = index visit with Prior 1; +2 Priors = index visit with Priors 1 and 2; +3 Priors = index visit with Priors 1, 2, and 3; multi-modal +1 Prior = index visit with Prior 1 for imaging and clinical priors from first and recent visit; multi-modal +2 Priors = index visit with Priors 1-2 for imaging and clinical priors from first and recent visit; multi-modal +3 Priors = index visit with Priors 1-3 for imaging and clinical priors from first and recent visit;

**Table S3:** Predicting high risk of prostate cancer within 5 years of the index visit: Performance comparison of risk refinement models. For each index visit during which imaging was performed, we identify test cases with at least three prior MRI exams for the image-based risk refinement model and with a recent PSA test (within 6 months of the index visit) and three prior PSA tests for the clinical risk refinement model. For each model, the best performance is highlighted in bold. For Clinical Only, the three prior visits (Priors 1, 2, and 3) were selected such that Prior 1 was the most recent visit (average time interval from index visit = 7 ± 6 months), Prior 3 was the initial visit (47 ± 25 months), and Prior 2 was the midpoint (25 ± 14 months). For Imaging Only, the three prior exams were taken from three consecutive prior imaging visits, with average intervals between index visit and prior visit of 2.1 ± 1.2 years for Prior 1, 3.3 ± 1.3 years for Prior 2, and 4.6 ± 1.4 years for Prior 3. Note that the total number of cases is smaller for 5-year risk than for current risk, as we only used the subset of the test cases with outcome assessments five years after the index visit.

| | False negatives | True positives | False positives | True negatives | Sensitivity | Specificity | AUC |
|---|---|---|---|---|---|---|---|
| Clinical data-based starting risk refined with clinical priors ("Clinical Only," n=89, positive =56) | | | | | | | |
| No prior | 3 | 53 | 24 | 9 | 0.95 | 0.27 | 0.86 |
| +1 Prior | 3 | 53 | 18 | 15 | 0.95 | 0.45 | 0.86 |
| +2 Priors | 3 | 53 | 17 | 16 | 0.95 | 0.48 | 0.87 |
| +3 Priors | 3 | 53 | **16** | 17 | 0.95 | 0.52 | 0.86 |
| Image-based starting risk refined with imaging priors only ("Imaging Only," n=28, positive=17) | | | | | | | |
| No prior | 2 | 15 | 7 | 4 | 0.88 | 0.36 | 0.88 |
| +1 Prior | 2 | 15 | 5 | 6 | 0.88 | 0.54 | 0.91 |
| +2 Priors | 2 | 15 | 3 | 8 | 0.88 | 0.73 | 0.93 |
| +3 Priors | 2 | 15 | **1** | 10 | 0.88 | 0.91 | 0.91 |
| Image-based starting risk refined with imaging and clinical priors (whenever available) ("Imaging + Clinical," n=28, positive=17) | | | | | | | |
| No prior | 2 | 15 | 7 | 4 | 0.88 | 0.36 | 0.88 |
| +1 Prior | 2 | 15 | 5 | 6 | 0.88 | 0.54 | 0.92 |
| +2 Priors | 2 | 15 | 2 | 9 | 0.88 | 0.82 | 0.93 |
| +3 Priors | 2 | 15 | **1** | 10 | 0.88 | 0.91 | 0.92 |

AUC = area-under-receiver-operating-characteristic curves; No prior = risk prediction with index visits only; +1 Prior = index visit with Prior 1; +2 Priors = index visit with Priors 1 and 2; +3 Priors = index visit with Priors 1, 2, and 3; multi-modal +1 Prior = index visit with Prior 1 for imaging and clinical priors from first and recent visit; multi-modal +2 Priors = index visit with Priors 1-2 for imaging and clinical priors from first and recent visit; multi-modal +3 Priors = index visit with Priors 1-3 for imaging and clinical priors from first and recent visit;

**Table S4:** Predicting high risk of developing prostate cancer within 5 years of the index visit with the Imaging Only risk refinement model. For each index visit during which imaging was performed, we identify test cases with at least three prior MRI exams for the image-based risk refinement model. The three prior exams were taken from three consecutive prior imaging visits, with average intervals between index visit and prior visit of 2.1 ± 1.2 years for Prior 1, 3.3 ± 1.3 years for Prior 2, and 4.6 ± 1.4 years for Prior 3. Note that, here we only consider index visits that were assessed as negative at the time of the visit.

|  | False negatives | True positives | False positives | True negatives | Sensitivity | Specificity |
|---|---|---|---|---|---|---|
| Image-based starting risk refined with at least three imaging priors ("Imaging Only," n=14, positive=3) | | | | | | |
| No prior | 0 | 3 | 7 | 4 | 1.0 | 0.36 |
| +1 Prior | 0 | 3 | 5 | 6 | 1.0 | 0.54 |
| +2 Priors | 0 | 3 | 3 | 8 | 1.0 | 0.73 |
| +3 Priors | 0 | 3 | **1** | **10** | 1.0 | 0.91 |

No prior = risk prediction with index visits only; +1 Prior = index visit with Prior 1; +2 Priors = index visit with Priors 1 and 2; +3 Priors = index visit with Priors 1, 2, and 3

**Table S5:** Effect of time interval between index visit and prior visit on predicting current risk of prostate cancer with temporal context: Performance comparison of risk refinement models on a subset of the test set with at least three prior visits. For each imaging-based index visit, we identify cases that have at least three prior MRI exams for the image-based risk refinement model and have a recent PSA (within 6 months of the index visit) with three prior PSA tests for the clinical risk refinement model. Here, we show the performance of the model when a single prior from different prior visits is integrated into our risk refinement framework. The average time to follow-up (years) between MR images from current and prior visits was 2.1 ± 1.2y for the recent prior, 3.3 ± 1.3y for the older prior, and 4.6 ± 1.4y for the oldest prior. The average time to follow-up (in months) between clinical data from current and prior visits was 7 ± 6m for the recent prior, 25 ± 14m for the older prior, and 47 ± 25m for the oldest prior. For each model, the best performance is highlighted in bold.

|  | False negatives | True positives | False positives | True negatives | Sensitivity | Specificity | AUC |
|---|---|---|---|---|---|---|---|
| Image-based starting risk refined with three imaging priors (n=135, positive=14) | | | | | | | |
| No prior | 1 | 13 | 61 | 60 | 0.93 | 0.49 | 0.86 |
| +Recent | 1 | 13 | 56 | 65 | 0.93 | 0.54 | 0.86 |
| +Older | 2 | 12 | 51 | 70 | 0.86 | 0.58 | 0.88 |
| +Oldest | 2 | 12 | **46** | 75 | 0.86 | 0.62 | 0.85 |
| Clinical data-based starting risk refined with three clinical priors (n=279, positive =44) | | | | | | | |
| No prior | 3 | 41 | 176 | 59 | 0.93 | 0.25 | 0.81 |
| +Recent | 3 | 41 | 175 | 60 | 0.93 | 0.25 | 0.82 |
| +Older | 3 | 41 | 156 | 79 | 0.93 | 0.34 | 0.83 |
| +Oldest | 3 | 41 | **140** | 95 | 0.93 | 0.40 | 0.85 |

AUC = area-under-receiver-operating-characteristic curves; No prior = risk prediction using the index visits only; +Recent = risk prediction using the index visit and a recent prior; + Older = risk prediction using the index visit and an older prior; +Oldest = risk prediction using the index visit and the oldest prior

**Table S6:** Acquisition parameters for the abbreviated bi-parametric prostate MR imaging (bpMRI) protocol used at our institution. The bpMRI protocol consists of axial T2-weighted imaging (T2WI) and diffusion-weighted imaging (DWI). The Echo-planar imaging sequence for DWI uses tri-directional diffusion sensitizing gradients with b-values of 50 s/mm$^2$ (b50) and 1000 s/mm$^2$ (b1000). The acquired b50 and b1000 data are used to calculate apparent diffusion coefficient (ADC) maps and b=1500 s/mm$^2$ (b1500) data used in this work.

| Acquisition Parameters | T2WI | DWI |
| --- | --- | --- |
| TE (ms) | 100 | 77 |
| TR (s) | 3.5-7.2 | 5.0-7.3 |
| Echo train length | 25 | 75 |
| In-plane resolution (mm$^2$) | 0.56x0.56 | 2.0x2.0 |
| Slice thickness (mm) | 3 | 3 |
| Matrix size | 320x320 | 100x100 |
| Field of view (mm) | 180x180 | 200x200 |
| NEX | 3 | b50: 4, b1000: 12 |

TE = echo time; TR = repetition time; NEX = number of averages